# Integration of a Balanced Virtual Manikin in a Virtual Reality Platform aimed at Virtual Prototyping

Antoine Rennuit[123], Alain Micaelli[1], Xavier Merlhiot[1], Claude Andriot[1], François Guillaume[2], Nicolas Chevassus[2], Damien Chablat[3], Patrick Chedmail[3]

(1) : CEA\LIST
Fontenay-aux-Roses, France
+33 1 46 57 70 80
E-mail : {rennuita, micaellia, merlhiotx, andriotc}@zoe.cea.fr

(2) : EADS\CCR
Suresnes, France
+33 1 46 97 30 00
E-mail : {antoine.rennuit, francois.guillaume, nicolas.chevassus}@eads.net

(3) : IRCCyN
Nantes, France
+33 2 40 37 69 33
E-mail : {Antoine.Rennuit, Damien.Chablat, Patrick.Chedmail}@irccyn.ec-nantes.fr

**Abstract:** the work presented here is aimed at introducing a virtual human controller in a virtual prototyping framework. After a brief introduction describing the problem solved in the paper, we describe the interest as for digital humans in the context of concurrent engineering. This leads us to draw a control architecture enabling to drive virtual humans in a real-time immersed way, and to interact with the product, through motion capture. Unfortunately, we show this control scheme can lead to unfeasible movements because of the lack of balance control. Introducing such a controller is a problem that was never addressed in the context of real-time. We propose an implementation of a balance controller, that we insert into the previously described control scheme. Next section is dedicated to show the results we obtained. Finally, we propose a virtual reality platform into which the digital character controller is integrated.

## 1- Introduction

Virtual humans are getting more and more important in the designing process, we show their usefulness below. But they are still painful to use. This paper introduces new control methods in a Virtual Reality (VR) platform aimed at virtual prototyping, that eases the use of digital characters.
We describe a control architecture that makes it able to interact with the product, enforce physical constraints such as joint limits, or anti-collision, and provides tools to make designers' job easier in an immersive real-time way. The main purpose of the paper is the introduction of balance control into this control architecture.
Real people movements are naturally balanced. But if not regarded, the movements produced by a computer do not respect balance in the general case. Hence generating unfeasible movements. The problem is shown on Figure 1.

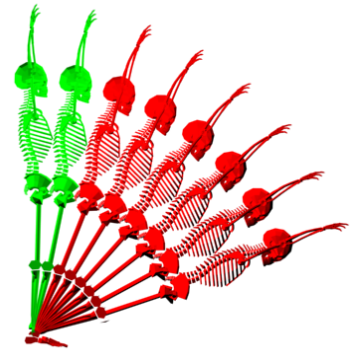

**Figure 1 : range of computer generated postures. The red skeletons describe postures which could not be held without falling. We want to avoid such situations which correspond to unfeasible postures.**

The balance controller introduced in 4- is to avoid balance loss.

In the remaining sections of the paper, we describe the interest of a good and easy to use animation controller. Then we detail the proposed control architecture. This architecture is shown to lack of balance control, thus we introduce it in another section. We show the results of the proposed controller. And finally, we show how it is integrated into a flexible virtual reality platform aimed at virtual prototyping.

## 2- Interest and context

The work we describe here takes place in a far larger context:

the industrial design process. The Virtual Human (VH) we are designing is to fit the concurrent engineering design approach. Large scale concurrent engineering (also known as "engineers' dream") is now regarded as very attractive [1]. It is held by Information Technology (IT) innovation as well as by new organisation and management methods. That is rational methods have to be developed to take advantage of industrial IT objects or machines, such as digital mock-up reviews to support large collaborative teams…

In such a framework, the Digital Mock-Up (DMU) is no longer an assembly model in a CAD tool, but an object managed by a Product Data Management (PDM) tool, which supports the product's integrity through collaborative work.

This collaborative aspect implies to control data flows : data exchange between people and IT machines, and data produced from other data thanks to people knowledge and software tools (knowledge management).

IT innovation for collective know how support (see Figure 2) is related to :

   - seamless virtual product simulation and analysis, from early to in service models (Virtual Prototyping (VP))

   - technical IT data flow from early investigation to downstream end users (data exchange)

   - knowledge cycles from early concepts to knowledge support of end user (capitalize, and restore knowledge)

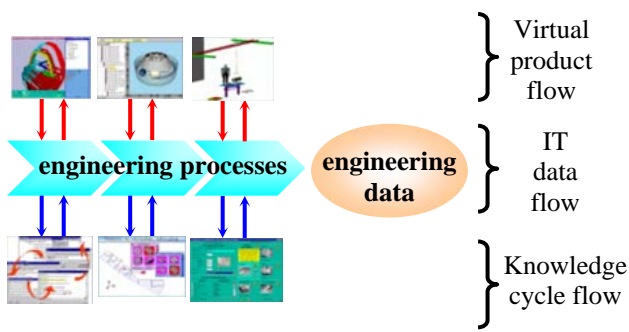

**Figure 2 : IT support of collective know how in engineering process**

Such a designing approach reduces time to market, development costs, design rework and retrofit, and heightens quality, safety, productivity of the designed product.

Virtual Prototyping's flow shown on Figure 2, is held by the digital mock-up, which enables to lead varied studies and tests (such as fluid or vibratory simulation, process planning…), in a digital way. Thus it makes it able to get rid of most physical mock-ups which are known to be expensive, and often obsolete before they can be used.

Unfortunately really complex simulations still cannot be undertaken through virtual prototypes. This is the case of problems related to human centric design. As it is a sore point when designing a product, digital humans were introduced into all major Product Lifecycle Management (PLM) tools such as in Delmia. But conducting studies in such frameworks is really long, painful, and expensive, mainly because of the lack of tools and functionalities associated to digital humans. This is due to the difficulty to tackle with, and control complex systems such as humans. The main problem is the one of control, which is rather poor (e.g. it takes a very long time to animate a character, controllers do not allow to interact with the product…). Moreover, this creates a risk for the designing team to botch human centric design, risk which must absolutely be avoided.

Introducing an efficient virtual human animation solution could help in all PLM's steps (see Figure 3).

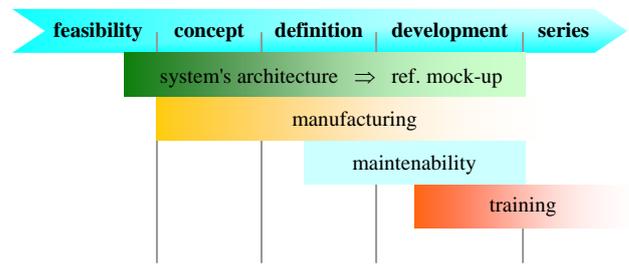

**Figure 3 : PLM in the concurrent engineering process, as seen by EADS (simplified scheme highlighting the steps that will benefit most of VH).**

Next we list, the contributions to the design process a good animation package would bring in each step of the PLM:

**VH for manufacturing**
*workcell layout :* machines and equipments positions to optimise cycle times and avoid hazards
*workflow simulation :* design manufacturing processes to eliminate inefficiencies and ensure optimal productivity. Simulate capabilities and limitations of humans to optimize the process
*reaching and grasping :* check if workers can access parts, equipments, and manipulate the tools needed to perform the task
*safety analysis :* ensure tasks are performed in a safe way
*strength analysis :* check if manipulating the product does not need extraordinary efforts, or create the potential for injuries
*energy expenditure :* calculate energy expended over time as workers perform a repetitive task, and optimise movement

**VH for training**
*manufacturing training :* use VR to train assembly workers on the virtual shop. Ability to modify reality to strengthen learning (ex : hide the blinding flash of lighting, when training welding, in order to see what we are doing).
*maintenance training :* leverage computer technology to train maintenance personnel from multiple locations

**VH for operating**
*positioning and comfort :* optimise user comfort, visibility, access to controls
*visibility :* ensure differently sized people see what is important when manipulating the product
*accessibility :* verify if the target population can easily climb in and out of the vehicle or equipment
*reaching and grasping :* test if controls are placed in such a way that everybody can operate them, also consider foot-pedal operations
*multi-persons interaction :* does the product fit collaborative work constraints?
*emergency situations :* check evacuation, and crowd movements in case of emergency *strength assessment :* check if operating the product does not need extraordinary force, or create the potential for injury

**VH for maintainability**
*reaching and grasping :* check if there is enough room for technicians to perform maintenance tasks, including space for tools
*part removal and replacement :* ensure that all technicians can efficiently install and remove parts
*visibility :* foresee what technicians can see when they perform a task
*strength capability :* ensure it is possible, and not too difficult for a technician to perform its task. Reveal the need for collaborative work when needed
*safety analysis :* be sure the technicians work in a safe environment

In the present paper we propose an architecture which aims at filling the gap between current possibilities of virtual humans, and expected performances - that would allow to bring the VH contributions enumerated above for each PLM steps. We think this architecture brings ease of use of digital characters. As explained in [6], we focus our work on interactive virtual humans enabling to drive avatars in real time thanks to motion capture devices, in a way that is shown to be very handy, easy to use (unfortunately this easiness is encountered only when using the system, not when developing it), and very persuasive for product reviewers or clients.

The control scheme we propose makes it able to interact with the product, to enforce physical limits (such as joint limits, or anti-collision), provides tools to ease the actor's job…

Thus we proposed the architecture given Figure 4, which distinguishes, and splits the simulation from the control itself. This architecture, which we introduced in [2], is innovative for the virtual reality domain, and was borrowed from robotics.

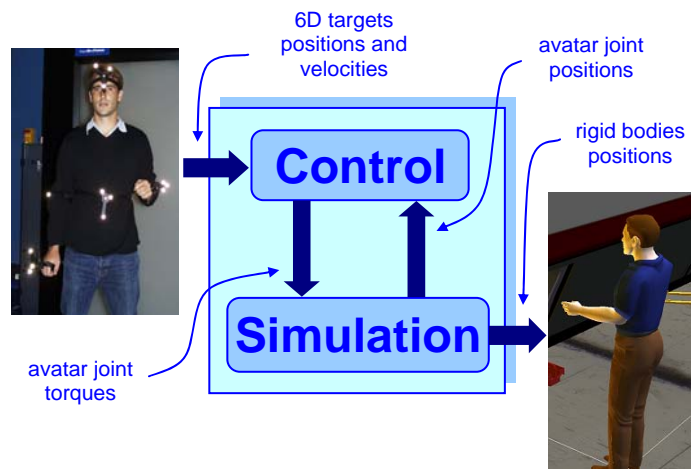

Figure 4 : Global scheme of the proposed system

Though efficient, this architecture can lead to unphysical movements of the avatars. This means the movements generated by the controller can be unfeasible… The most striking problem turns out to be the loss of balance in some situations. Of course real humans movements are always balanced, this is a physical constraint. As stated above, the goal of this paper is to enforce balance thanks to a balance controller.

## 3- Detailed control architecture

The proposed architecture we developed can be seen in a detailed way on the following scheme.

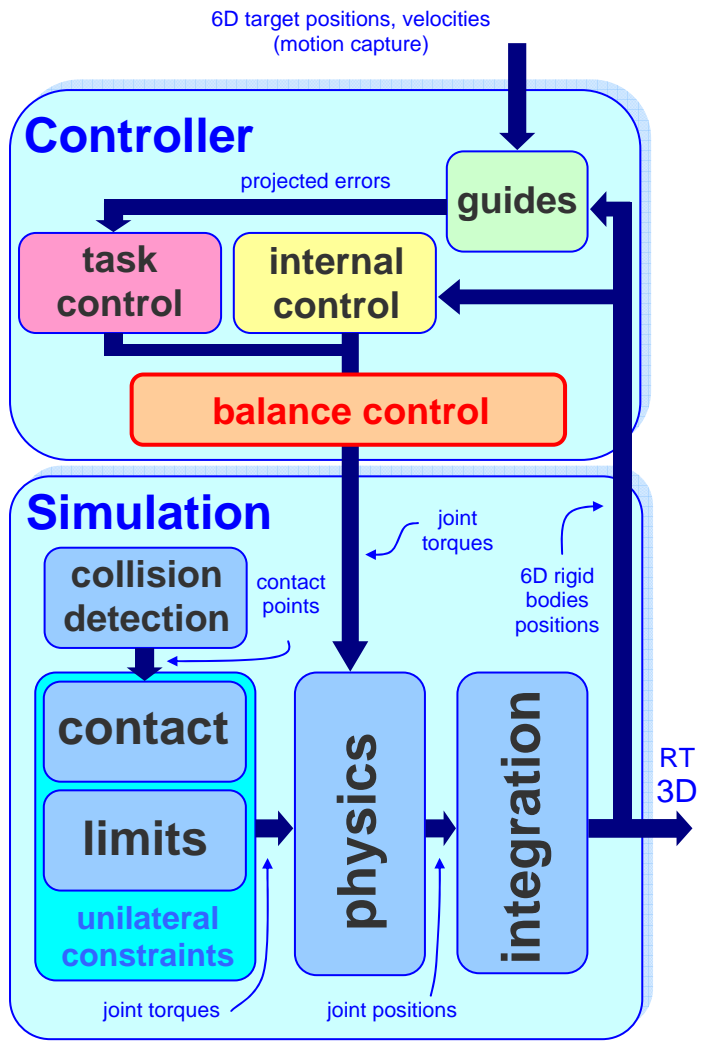

Figure 5 : Detailed architecture

This control scheme distinguishes two main blocks: on the one hand we find the simulation, which emulates the real world's physical laws and the manikin, and on the other hand the controller drives the virtual human to the desired goal, expressed in terms of task space targets positions, and other constraints such as balance (which is innovative into a real-time framework - its introduction into the control scheme is detailed in 4-), guides…. Now we rapidly detail the internal behavior of the control scheme.

Desired targets positions are received from a motion capture device. The error is projected in a passive way, thanks to mechanical analogies as explained in [2] (in the general case, projections break passivity – a notion useful when studying stability -, thus we use a virtual mechanical system - which is known to be passive – to restrict, in a virtual and passive way, the movements of the motion capture targets), hence creating virtual guides (in green on Figure 5), an example of these guides in action can be seen on Figure 11. Then the projected error goes through a task space corrector (in pink), which will generate a compensation for the error. As virtual humans are highly redundant systems, we can add to the task space control an internal control, which must not interfere with task space control (in yellow).

Now comes the management of unilateral constraints. As seen in 4-, balance control (in red) can be seen as a unilateral constraint enforcing the balance of the virtual human being constrained (balance control was not addressed so far in real-time immersive controllers, we express the need for such control, and solve it in 4-). Joint limits and contact response are managed in the Simulation block, because they are not specific to character animation. They respectively enforce joint limits (of course). The non penetration with environment. And also the interaction with environment: that is virtual humans can apply forces onto the virtual environment, hitting, pushing, and pulling as a real human would do on a real environment...

We use GVM, and LMD++, two packages developed by CEA\LIST, to perform physical simulation [4].

## 4- Balance control

### 4.1- Problem and state of the art

Now we show the need for balance control.

Let's take an example of a giant actor which motion is retargeted on a dwarf avatar. When performing its movements, the actor can move the targets in such positions that kinematically (or geometrically) speaking the digital human can reach the targets, but if it does so, its balance is not enforced anymore, thus leading to unfeasible movements of the avatar. This is what happens on the following scheme:

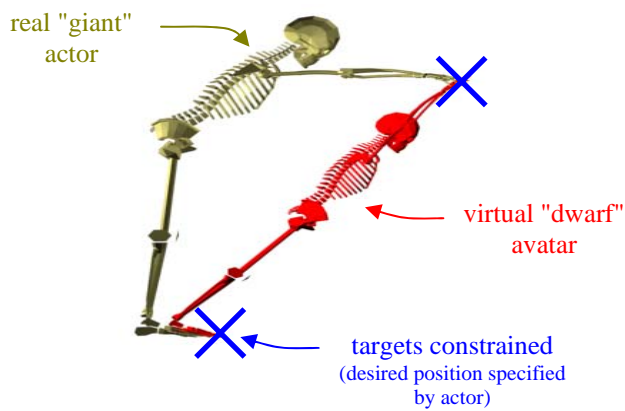

**Figure 6 When retargeted, the giant's motion gives an unbalanced movement on the dwarf**

To account for such situations, Laszlo, van de Panne, and Fiume introduced Limit Cycle Control [7] which makes it able to compensate for small disturbances of a cycling movement (seen in a reduced state space). In the given example, the compensation for regulation variables called *up-vector* is made through control variables that are chosen to be the hip pitch and roll angle. This approach is interesting for cycling movements such as walking, but we want to be able to perform whatever needed movement, without restriction. Hodgins and Wooten [8], proposed to control ankles and hips angles to enforce static equilibrium, but their approach is specific to one kind of movements (vaulting in their example). Faloutsos, van de Panne, and Terzopoulos [9] detect disturbances thanks to the famous notion of *support polygon*, and try to balance their virtual manikin thanks to the only ankle's stiffness correction. Although their approach tackles particularly well with situations where balance is lost, their control is rather restrictive.

When an actual human's balance is disturbed, the natural reaction is not to adjust a single or a couple of joints, the whole body is involved in the balance recovering.

Liu, and Popovic [10], proposed to blend a distance to balance in the objective of a space-time optimization (ensuring an answer to disturbance distributed on all joints). This enables the generation of nice movements from sketches. Unfortunately, because of the optimization that is made on the whole motion at once (this technique is called space-time optimization), the method is unuseful to our purpose: the entire movement must be known before retargeting. Fang and Pollard brought a physical filter [11], which does not allow real time performances, though it is much faster than previous attempts.

The biped robots community encounters the same kind of balance problem as we do. Besides the solutions already shown they also use the well known notion of Zero Moment Point (ZMP) [12], which allows to study dynamic equilibrium, but only when all contacts with environment occur on a plane (e.g. walking, running…). Harada et al. [13] extend the notion of ZMP to situations where contacts are not located on the same plane anymore, and ensure balance thanks to a method they propose, based on linear complementarity.

The virtual humans control scheme we propose is to support balance control in an interactive manner.

### 4.2- Proposed solution

Our main purpose was to check, thanks to a simple controller, if the architecture described in [2] could handle balance control. That is why we chose to build the balance controller on the well known concept of support polygon.

This notion states that *walking systems remain statically balanced so long as the vertical projection of their center of mass stays inside the convex hull of contact points* [14].

Balance can be seen as a unilateral constraint rather straightforwardly. The unilateral constraint solution is given by the resolution of a Linear Complementarity Problem (LCP). Its superiority with respect to regulation methods is detailed in [15].

A general LCP can be expressed as follows:

Given $q \in R^p$, $M$ a $p \times p$ matrix, $w \in R^p$, and $z \in R^p$, ensure unilateral constraints $\omega \geq 0$, and $z \geq 0$, knowing complementarity $\omega^T z = 0$, and the relation between $\omega$, and $z$: $\omega = Mz + q$.

Usually expressed in a shorter shape: $0 \leq \omega \perp z \geq 0$, with $\omega = Mz + q$. $\omega$ can be seen as a control variable, and $z$ as the distance to constraint, further explanations can be found in [16].

We now express the balance problem as a LCP. We will illustrate the approach on a virtual manikin standing straight, with both feet on the ground. We know the projected center of mass must lie inside the support polygon. As seen on Figure 7, we will approximate the support polygon of both feet by an ellipse, this is done without loss of generality, because the LCP could be expressed with a polygon as well.

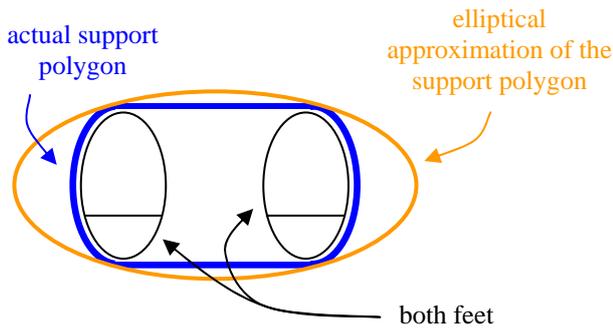

**Figure 7 : Elliptical approximation of the actual support polygon**

The support polygon approximation makes it able for us to regard the configuration of a virtual human as balanced when the vertical projection of the center of mass lies inside the elliptical limit, as seen on Figure 8:

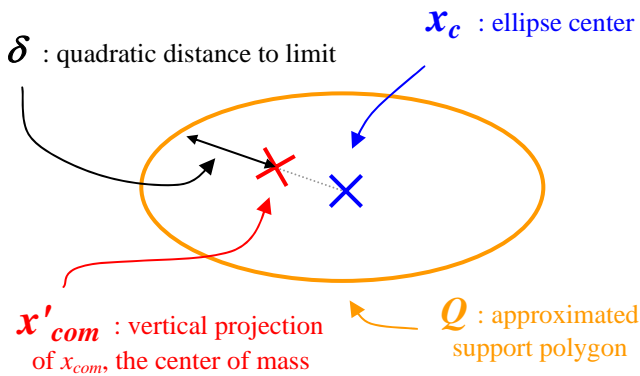

**Figure 8 : Schematic diagram of the balance control: we aim at keeping $x_{com}$ inside of Q.**

As for our problem, the LCP, is expressed as follows:
We want $\delta$, the quadratic distance to limit, to remain positive or zero, if this constraint breaks, the LCP solver, will modify $\Gamma_{LCP}$ (the joint torques due to the unilateral constraint enforcement) to enforce the constraint on $\delta$, this modification being done according to the relation between $\delta$ and $\Gamma_{LCP}$.

Reference [4] shows this relation between $\delta$ (the variable being constrained), and $\Gamma_{LCP}$ (the control variable), can be expressed by the Jacobian matrix of $\delta$, with respect to $q$ (the joint parameters of the virtual human), when the equation of evolution of the system (the dynamics equation) is known. That is the LCP solver's input will be the Jacobian matrix of $\delta$. Thus we now express $\delta$, and its Jacobian matrix.
As $Q$, is an ellipse, $\delta$ can be written as:

$$\delta = d^2 - \|P(x_{com} - x_c)\|_Q^2, \quad (1)$$

with $d$, the maximum distance, $P$, the vertical projection, and $Q$, the metric corresponding to the ellipse $Q$.
$J$, the Jacobian matrix of $\delta$, can be expressed by:

$$J = \frac{\partial \delta}{\partial q}, \quad (2)$$

thanks to eq. (1), $\partial \delta$ becomes:

$$\partial \delta = -\partial \|P(x_{com} - x_c)\|_Q^2, \quad (3)$$

considering the virtual human is not changing its support polygon (that is double feet support, is made independent of single foot support, during a walk), we have:

$$\partial \delta = -(x_{com} - x_c)^T P^T (Q + Q^T) \partial (P(x_{com} - x_c))$$
$$= -(x_{com} - x_c)^T P^T (Q + Q^T) P \partial x_{com}$$
$$= -(x_{com} - x_c)^T P^T (Q + Q^T) P J_{com} \partial q, \quad (4)$$

with $J_{com}$ being the Jacobien matrix of the center of mass. We know have to express $J_{com}$.

The position of the center of mass $p_{com(0)}$ of an articulated system, expressed in the base frame $0$, is given by:

$$p_{com(0)} = \frac{\sum m_i p_{com\_i(0)}}{\sum m_i}, \quad (5)$$

with $p_{com\_i(0)}$ being the position of the center of mass of the $i^{th}$ solid, thus the velocity $v_{com/0(0)}$ of the full system's center of mass seen from base frame $0$, and expressed in the same frame is:

$$v_{com/0(0)} = \frac{\partial p_{com(0)}}{\partial t} = \frac{\sum m_i \frac{\partial p_{com\_i(0)}}{\partial t}}{\sum m_i}$$
$$= \frac{\sum m_i v_{com\_i/0(0)}}{\sum m_i}$$

so $J^r_{com/0(0)}$ (written $J_{com}$ in eq. (4)), the Jacobien matrix of $v_{com/0(0)}$ is given by:

$$J^r_{com/0(0)} = \frac{\sum m_i J^r_{com\_i/0(0)}}{\sum m_i}, \quad (6)$$

$J^r_{com/0(0)}$ is a reduced Jacobien matrix, because a center of mass position has 3 translational components, whereas a full solid position is 6D (3 rotations more), the Jacobien matrix associated to a general solid position is written $J_{A \in i/j(k)}$ (one must understand Jacobien of the speed of point $A$ fixed in frame $i$, in its movement with respect to frame $j$, expressed in base $k$). The relation between $J_{A \in i/j(k)}$, and $J^r_{A \in i/j(k)}$ is given by:

$$J^r_{A \in i/j(k)} = (0_{3*3} \quad I_3) J_{A \in i/j(k)}$$
$$= S J_{A \in i/j(k)} \quad (7)$$

Knowing this, we can now express $J^r_{com/0(0)}$, thanks to the Jacobien matrices of the center of mass of each solid of the articulated system $J_{com\_i/0(0)}$. Eq. (6) gives us:

$$J^r_{com/0(0)} = \frac{\sum m_i J^r_{com\_i/0(0)}}{\sum m_i}$$
$$J_{com} = J^r_{com/0(0)} = \frac{\sum m_i S J_{com\_i/0(0)}}{\sum m_i} \quad (8)$$

Introducing $\delta$, and $J_{com}$, into the LCP solver (which choice is out of scope) gives the interesting results seen in next section.

## 5- Results

The balance controller we designed is aimed at enforcing static equations, in a framework allowing real time animation, interaction with environment, virtual guides... Its great behavior is illustrated bellow, first we show figures of the system's behavior to collision:

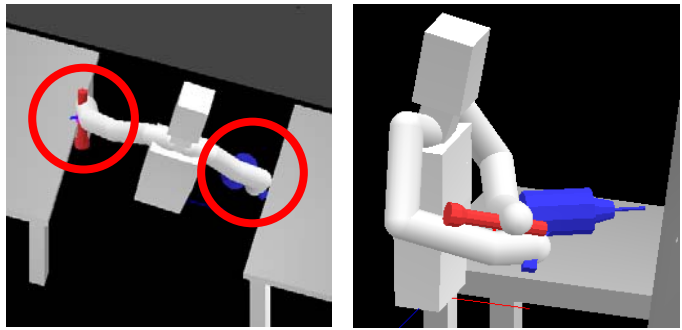

**Figure 9 : Double and self collision**

On the curve Figure 10, we can see the height of the table, which must not be penetrated (orange), and the height of the virtual human's hand (green), while reaching, and leaning on the table. We see that the hand never penetrates the table.

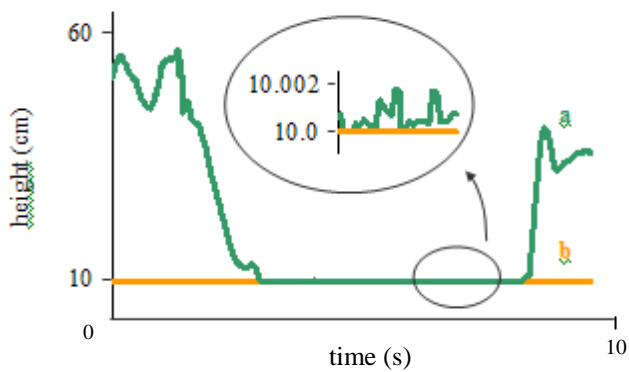

**Figure 10 : Hand (a), and obstacle's (b) height: no penetration.**

We now test the virtual guides approach we have implemented. The experiment consists in drilling a hole in a wall thanks to a drill, while lighting the future hole location thanks to a hand light. The drill can only move along a fixed axis with a fixed orientation. This means that the controller leaves only one degree of freedom to the operator. The direction of the spotlight is also driven automatically (leaving the three degrees of freedom of the light's position to the operator). Figure 11 depicts the ideal axis in green (a) and actual axis are in red (b).

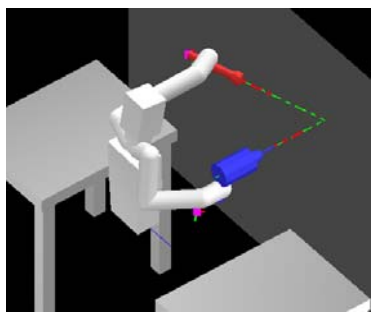

**Figure 11 : Worker drilling a hole, guided by virtual mechanisms**

In order to see the efficiency of the method, we drew the angle between the ideal axis, and the actual axis of the drill, as seen on Figure 12; in the case where the operator is completely free (green), and in case where the guide is on (orange).

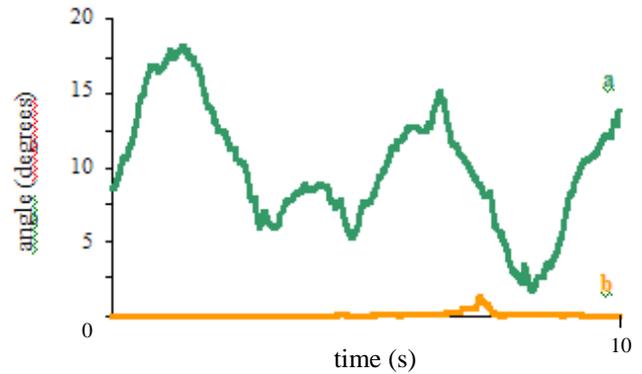

**Figure 12 : Angle between ideal and actual axis of the drill, (a) without guide, and (b) with guide.**

Now we show the balance controller's action:

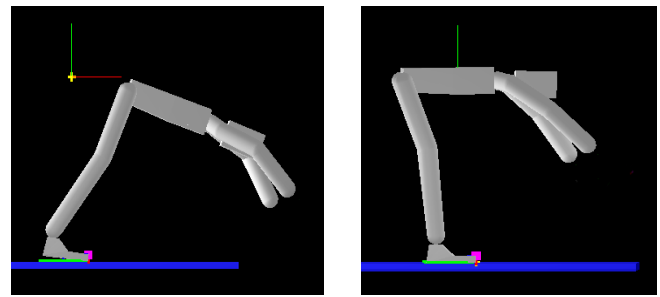

**Figure 13 : Unbalanced dwarf being controlled (left image), and the same dwarf being balance controlled, while the giant actor performs the same movement (right image).**

We can see that the configuration proposed by the unbalanced controller is unfeasible, whereas with be balance controller on, the system behaves well: a real human could adopt this posture without falling.

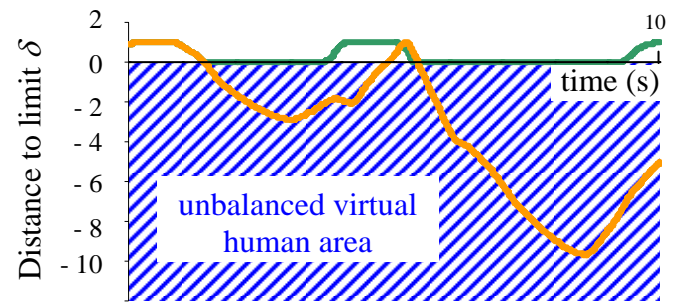

**Figure 14 : Distance to limit in the unbalanced case (orange), and in the balanced case (green). The distance to limit is expressed in multiples of $d$.**

Figure 14 shows the constraint is sharply enforced in case of balance control (green line), which is not the case when no control is done on balance (orange line).
We should notice that the model we chose as for balance (observing the projection of the center of mass), can only enforce static balance of walking avatars, that is the balance model is correct so long as the manikin does not interact with environment in another way than with the feet, on a plane surface.

The interaction of the digital human with the environment in a real-time immersive framework is an innovation in itself, presented in [2]. But the main issue and the innovative feature introduced in this paper is the possibility added to control balance in an automated way while retargeting the movement. As stated above, the main purpose of this controller is to enforce feasibility of the obtained movement.

## 6- Integration in a virtual reality platform

The virtual framework we describe here is to bring modularity, flexibility, and ease of use – of course.
The whole framework is clustered since physics, motion capture, and rendering, for example, are distributed. It was a choice made on purpose enabling to rely on light material, to heighten flexibility. The idea is when one want to add a function to the platform, we just have to add a PC to handle the additional computation load.
Unfortunately the addition of new functionalities to the architecture cannot be done as is, the different softwares that we use are not designed to work together. Thus our main work is integration, to make things work together.

As stated above, we try to concentrate on integration instead of development which is much more involving. Thus we try to rely on middleware game applications (the game industry's technologies are very close to the technologies used in virtual reality), and/or open source packages. Nevertheless, every needed functionalities are not already addressed. In [6] we expressed the need as for virtual humans. We could not find satisfying solutions, thus we decided to design our own virtual human controller. This paper is the outcome of it.
The virtual reality platform we are using comprises five main elements.

**Simulation**
A real-time physical simulator. Such simulators allow to describe the virtual world, and the objects contained in it. The various elements of the virtual scene have basic attributes (shape, location in space…), and complementary others that enrich the realism or provide specific information on status (a color coding to indicate a heat level for instance). The simulators that are used are enriched with a graphical layer to enhance realism with textures, lighting, camera movements. It can also be noted that the geometry for physics, is usually simplified with respect to the geometry that is displayed (the display geometry being usually too complicated to be handled efficiently by the physics module).

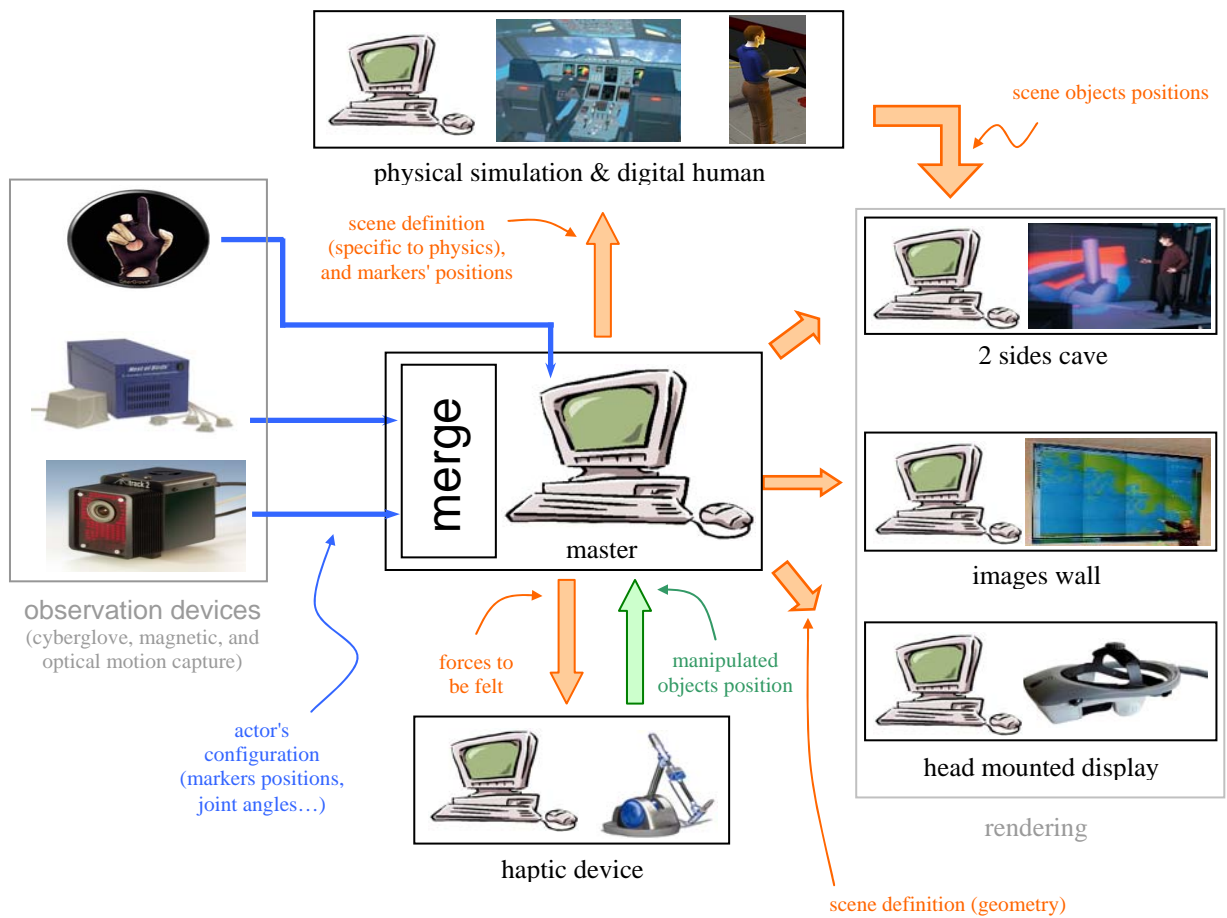

**Figure 15 : Our distributed virtual reality platform. All display computers know the scene's geometry, objects positions are updated by simulation.**

In an actual scene, a number of physical processes are involved depending on what we wish to simulate, be they thermal, wave, mechanical, dynamical, etc… Depending on the nature of these processes, the model equations used, and resolution methods will vary.

At the moment we mainly use dynamical simulators enabling to emulate the dynamics of rigid bodies, mechanical joints, joints limits, contact with friction…

We concentrate our study on the integration scheme that is implemented in the simulator, because it determines the whole system's stability. And also on constraints solvers which has a great influence on the resulting movements' quality.

**Motion capture**

Real-time motion-capture systems are used to observe the actor's movements, in order to retarget them onto virtual objects, and animate the digital character.

We use two types of motion capture systems: optical and magnetic. The optical device is used for precision and ease of use. It comprises 4 high range cameras, that make it able to have a wide observing volume. These targets are passive, avoiding the need for cumbersome instrumentation on the moving object (quick positioning, no electrical leads). Lighting in the near infrared is associated to each camera. Unfortunately optical systems suffer from the occlusion problem (the optical markers that are observed must lie in the optical range of cameras, that is if the character hides the markers, information is lost). In order to handle such situations, we also use a magnetic device which does not suffer from this problem, but the quality of the observation is not as good, and is very sensitive to metal proximity. So to take the best out of both observation methods we merge both sources.

**Haptic feedback**

Haptic devices are added to this framework. We use 6D interfaces. These devices provide you with 6 degrees of freedom, and also force and torque feedback. These haptic interfaces enable their users to touch and handle objects located in the virtual environment, with high sensitivity. It offers many prospects for our designing sector for virtual prototyping, because they enable to give haptic sensation back…

**Graphic display**

A cluster of PCs is dedicated to graphics. Allowing to display as many views as needed, each PC handling a point of view. This is particularly useful in case of project reviews: a general point of view of the product being designed is displayed for all reviewers on an images wall, or on a cave. Stereovision can be enabled, in such cases all reviewers are given active stereo glasses. The image wall can display as many views as needed, this is very flexible, thus another PC can compute another interesting view which will be displayed in a screen's corner. It can also be very interesting to compute a view for an actor being immersed: the actor is to drive a virtual human, and therefore should see what the virtual human would see if it was real. In order to give a better immersion feeling, and reach the *presence* sensation described in [5], we provide the immersed actor with a stereoscopic head mounted 3D display system.

On must pay particular attention to the keenness of vision, the field of view, the contrast, and the colour range enabled by head mounted display. These characteristics are critical, they determine the quality of the display, and makes it able to reach presence.

## 7- Conclusion

In this paper, we first briefly described the problem to be solved then we reminded the interest of digital humans in the context of concurrent engineering. We thus described a control architecture aimed at driving virtual humans in a real-time immersive manner. This control scheme makes it able to interact with the product through motion capture. Unfortunately, we showed such controllers could produce - in the general case – unfeasible movements because of the lack of balance control. Introducing such a control was never addressed before in such a real-time context. We proposed an implementation of such a controller that we inserted into the described architecture. Then we showed the obtained results, and expressed the innovation carried by the present paper. Finally we proposed a virtual reality platform into which we integrated the animation controller.

Our main purpose was to validate the possibility to balance virtual humans on the control architecture described in [2], thanks to a simple balance model.

Hence the natural continuation will be to extend the balance controller possibilities to handle multiple contacts with friction.

## 8- References

**[1]** Michel Dureigne, Serge Tichkiewitch, Daniel Brissaud, "Collaborative large engineering : from IT dream to reality", Kluwer Academic Publishers, 2004.

**[2]** A. Rennuit, A. Micaelli, X. Merlhiot, C. Andriot, F. Guillaume, N. Chevassus, D. Chablat, P. Chedmail, "Passive Control Architecture for Virtual Humans", submitted to IEEE/RSJ International Conference on Intelligent Robots and Systems IROS'2005, august 2005, Edmonton, Canada.

**[3]** A. Rennuit, A. Micaelli, X. Merlhiot, C. Andriot, F. Guillaume, N. Chevassus, D. Chablat, P. Chedmail, "Balanced Virtual Humans Interacting with their Environment", submitted to SCSC2005, Philadelphia, USA.

**[4]** X. Merlhiot, "Dossier d'analyse des algorithmes embarqués", CEA Technical Report, January 2005, to be published in the framework of the LHIR RNTL project.

**[5]** D.R. Mestre, "Interactions entre réalité virtuelle et neurosciences comportementales", Proceedings of Virtual Concept 2002, Biarritz France, 9-10 October 2002.

**[6]** A. Rennuit, A. Micaelli, C. Andriot, F. Guillaume, N. Chevassus, D. Chablat, P. Chedmail, "Designing a Virtual Manikin Animation Framework Aimed at Virtual Prototyping.", IEEE VRIC 2004.

**[7]** J. Laszlo, M. van de Panne, E. Fiume, "Limit Cycle Control, and its Application to the Animation of Balancing and Walking", Proceedings of ACM SIGGRAPH, New Orleans, Louisiana, August 1996.

**[8]** J.K. Hodgins, W.L. Wooten, "Animating Human Athletes", in Robotics Research: The Eighth International Symposium, 1998. Y. Shirai and S. Hirose (eds). Springer-Verlag: Berlin, 356-367.

**[9]** P. Faloutsos, M. van de Panne, D. Terzopoulos, "The Virtual Stuntman: Dynamic Characters with a Repertoire of Autonomous Motor Skills", Computers and Graphics, Volume 25, Issue 6, December, 2001, pp. 933-953.

**[10]** C.K. Liu, Z. Popovic, "Synthesis of Complex Dynamic Character Motion from Simple Animations", Proceedings of ACM SIGGRAPH, July 2002, San Antonio USA.

**[11]** A.C. Fang, N.S. Pollard, "Efficient Synthesis of Physically Valid Human Motion", Proceedings of ACM SIGGRAPH, July 2003, San Diego, USA.

**[12]** M. Vukobratovic, B. Borovac, "Zero Moment Point – Thirty Five Years of Its Life", International Journal of Humanoid Robotics, Vol. 1, No. 1 (2004) 157-173.

**[13]** K. Harada, H. Hirukawa, F Kanehiro, K. Fujiwara, K. Kaneko, S. Kajita, M. Nakamura, "Dynamical Balance of a Humanoid Robot Grasping an Environment", Proceedings of 2004 IEEE/RSJ International Conference on Intelligent Robots and Systems IROS'2004, October 2004, Sendai, Japan.

**[14]** P.B. Wieber, "On the Stability of Walking Systems", Proceedings of the International Workshop on Humanoid and Human Friendly Robotics, 2002.

**[15]** C. Duriez, "Contact frottant entre objets déformables dans des simulations temps-réel avec retour haptique", PhD thesis, Université d'Evry, December 2004.

**[16]** K.G. Murty, F.T. Yu, "Linear Complementarity, Linear and Nonlinear Programming", Heldermann Verlag, Berlin, 1988.